\documentclass[10pt,twocolumn,letterpaper]{article}

\usepackage{cvpr}
\usepackage{times}
\usepackage{epsfig}
\usepackage{graphicx}
\usepackage{multirow} 
\usepackage{amsmath}
\usepackage{amssymb}
\usepackage{bm} 
\usepackage{booktabs}
\usepackage[export]{adjustbox} 
\usepackage{IEEEtrantools} 
\usepackage{multirow} 
\usepackage{tabularx} 
\usepackage{caption} 
\usepackage[normalem]{ulem} 
\usepackage[title]{appendix}

\usepackage[pagebackref=true,breaklinks=true,letterpaper=true,colorlinks,bookmarks=false]{hyperref}

\newcommand{\eat}[1]{{}}

\newcommand{\bx}{\bm{x}}
\newcommand{\tx}{\tilde{\bm{x}}}
\newcommand{\bX}{\bm{X}}

\newcommand{\bq}{\bm{q}}

\newcommand{\bk}{\bm{k}}
\newcommand{\bK}{\bm{K}}
\newcommand{\bv}{\bm{v}}
\newcommand{\bV}{\bm{V}}

\newcommand{\bg}{\bm{g}}
\newcommand{\bG}{\bm{G}}
\newcommand{\bb}{\bm{b}}

\newcommand{\bw}{\bm{w}}
\newcommand{\bW}{\bm{W}}
\newcommand{\bz}{\bm{z}}
\newcommand{\bZ}{\bm{Z}}
\newcommand{\tz}{\tilde{\bm{z}}}
\newcommand{\bd}{\bm{d}}

\newcommand{\smallfigmoveup}{\vspace{-0.0mm}}
\newcommand{\secmoveup}{\vspace{-1.2mm}}
\newcommand{\postfigmoveup}{\vspace{-5.0mm}}

\newcommand{\capmoveup}{\vspace{-5.mm}}

\cvprfinalcopy 

\def\cvprPaperID{7245} 

\begin{document}

\title{Transform and Tell: Entity-Aware News Image Captioning}

\author{Alasdair Tran, Alexander Mathews, Lexing Xie\\
Australian National University\\
{\tt\small \{alasdair.tran,alex.mathews,lexing.xie\}@anu.edu.au}
}

\maketitle


\begin{abstract}
    We propose an end-to-end model which generates captions for images embedded
    in news articles. News images present two key challenges: they rely on
    real-world knowledge, especially about named entities; and they typically
    have linguistically rich captions that include uncommon words. We address
    the first challenge by associating words in the caption with faces and
    objects in the image, via a multi-modal, multi-head attention mechanism. We
    tackle the second challenge with a state-of-the-art transformer language
    model that uses byte-pair-encoding to generate captions as a sequence of
    word parts. On the GoodNews dataset~\cite{Biten2019GoodNews}, our model
    outperforms the previous state of the art by a factor of four in CIDEr
    score ($13 \rightarrow 54$). This performance gain comes from a unique
    combination of language models, word representation, image embeddings, face
    embeddings, object embeddings, and improvements in neural network design.
    We also introduce the NYTimes800k dataset which is 70\% larger than
    GoodNews, has higher article quality, and includes the locations of images 
    within articles as an additional contextual cue.
\end{abstract}



\secmoveup
\section{Introduction}

The Internet is home to a large number of images, many of which lack useful
captions. While a growing body of work has developed the capacity to narrate
the contents of generic images~\cite{Donahue2015LongTR, Vinyals2015ShowAT,
Fang2015FromCT, Karpathy2015DeepVA, Rennie2017SelfCriticalST, Lu2017KnowingWT,
Anderson2017BottomUpAT, Cornia2019ShowCT}, these techniques still have two
important weaknesses. The first weakness is in world knowledge. Most captioning
systems are aware of generic object categories but unaware of names and places.
Also generated captions are often inconsistent with commonsense knowledge. The
second weakness is in linguistic expressiveness. The community has observed
that generated captions tend to be shorter and less diverse than human-written
captions~\cite{vinyals2016show,li2018generating}. Most captioning systems rely
on a fixed vocabulary and cannot correctly place or spell new or rare words.
\eat{A growing body of work seeks to automatically generate captions that
describe the objects and relationships using only visual cues extracted from
the image itself~\cite{Donahue2015LongTR, Vinyals2015ShowAT, Fang2015FromCT,
Karpathy2015DeepVA, Rennie2017SelfCriticalST, Lu2017KnowingWT,
Anderson2017BottomUpAT, Cornia2019ShowCT}. While generic image descriptions
have their uses, such as for individuals with vision impairments, they are
often of less benefit to the average user. To produce more useful image
captions we need to go beyond generic descriptions and introduce information
that cannot be gleaned directly from the image alone. Fortunately, many images
have an associated context such as a news article, web page, or social media
post, which give the image greater meaning than can be extracted from its
pixels. To generate captions that go beyond generic description and actually
add information that could not be gleaned from the image alone we must take
this context into account. We focus on the news image captioning task in order
to design practical methods for exploiting contextual information. News image
captioning is an interesting instance of contextual captioning where news
articles provide context to images.}

\begin{figure}[t]
	\begin{center}
		\includegraphics[width=0.99\linewidth]{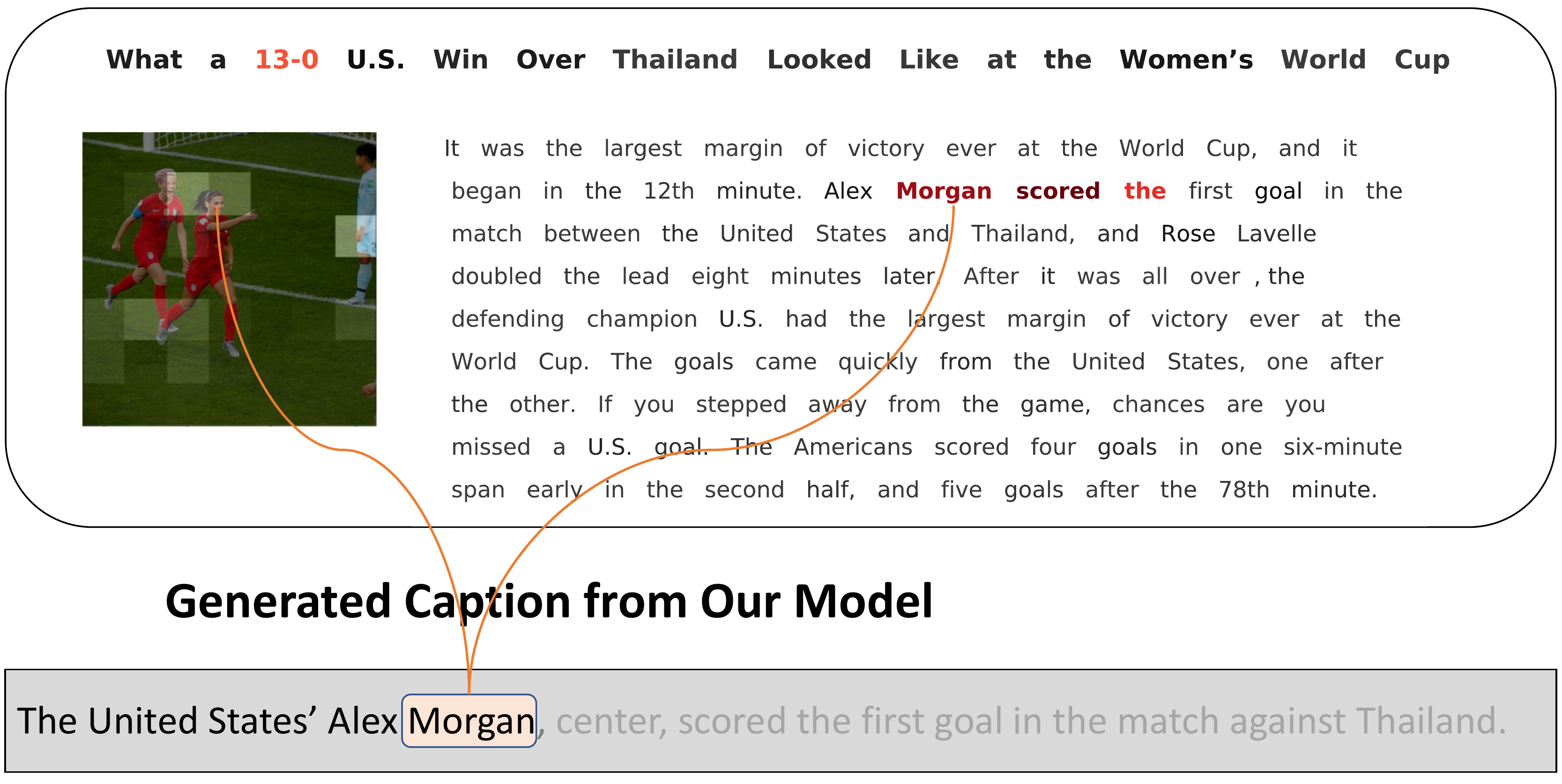}
	\end{center}
	\capmoveup
   \caption{An example of entity-aware news image captioning. Given a news
	 article and an image (top), our model generates a relevant caption
	 (bottom) by attending over the contexts. Here we show the attention scores
	 over the image patches and the article text as the decoder generates the
	 word ``Morgan". Image patches with higher attention have a lighter shade,
	 while highly-attended words are in red. The orange lines point to
	 the highly attended regions.}
	 \postfigmoveup
	\label{fig:teaser}
\end{figure}

\eat{However, images on the internet are
often associated with a context such as a news article, web page, or social
media post, which give the image greater meaning than can be extracted from its
pixels. To generate captions that go beyond generic description and actually
add information that could not be gleaned from the image alone we must take
this context into account.}

News image captioning is an interesting case study for tackling these two
challenges. Not only do news captions describe specific people, organizations
and places, but the associated news articles also provide rich contextual
information. The language used in news is evolving, with both the vocabulary
and style changing over time. Thus news captioning approaches need to adapt to
new words and concepts that emerge over a longer period of time (e.g. {\em
walkman} in the 1990s or {\em mp3 player} in the 2000s). Existing
approaches~\cite{Tariq2017ACE,Ramisa2016BreakingNewsAA,Biten2019GoodNews} rely
on text extraction or template filling, which prevents the results from being
linguistically richer than the template generator and are error-prone due to
the difficulty in ranking entities for gap filling. Successful strategies for
news image captioning can be generalized to images from domains
with other types of rich context, such as
web pages, social media posts, and user
comments.

\eat{Captions for news images, such as the example in Figure~\ref{fig:teaser},
typically contain details which cannot be derived from the image alone. They
also frequently contain proper nouns such as names of people, places, and
organizations -- in many cases these proper nouns are rare (most people and
places do not have many news articles written about them). A system capable of
generating high quality news captions should therefor make extensive use of the
provided context and be tuned for generating rare proper nouns. Existing
approaches to news image captioning~\cite{Tariq2017ACE,
Ramisa2016BreakingNewsAA,
	Biten2019GoodNews}  rely on text extraction or
template filling to deal with rare contextual terms such as names of people and
organizations. This makes them relatively inflexibility and means they cannot
be trained end-to-end. Moreover, existing approaches do not include specialized
visual models for frequent nouns -- experiments on the MSCOCO dataset have
shown
that pre-trained object detectors tuned for frequent nouns
lead to more accurate captions~\cite{Wu2016HighLevel,Gan2017Semantic}.}

\eat{The contextual word
	embeddings provided by these methods have helped establish new
	state-of-the-art
	results on many natural language understanding benchmarks including GLUE
	\cite{Wang2019GLUE}, SuperGLUE~\cite{Wang2019SuperGLUEAS}, and SQuAD
	\cite{Rajpurkar2016SQuAD, Rajpurkar2018KnowWY}, even surpassing the human
	baselines in many cases.}

\eat{In parallel to the development of these pre-training methods, many novel
techniques have been proposed to improve training convergence and handle more
diverse data inputs. The most significant contribution is the use of byte-pair
encoding (BPE)~\cite{Sennrich2015NeuralMT} to represent a rare word as a
sequence of subword units, thus giving models the ability to handle an open
vocabulary.}

\eat{that 1) combines specialized modules for
incorporating and selectively attending to image features, human faces, and
news article text and 2) applies a state-of-the-art sequence generation model
which is able to generate rare tokens, such as proper names, even when they do
not form part of the training data. Our model relies on a }

We propose an end-to-end model for news image captioning with a novel
combination of sequence-to-sequence neural networks, language representation
learning, and vision subsystems. In particular, we address the knowledge gap by
computing multi-head attention on the words in the article, along with faces
and objects that are extracted from the image. We address the linguistic gap
with a flexible byte-pair-encoding that can generate unseen words. We use
dynamic convolutions and mix different linguistic representation layers to make
the neural network representation richer. We also propose a new dataset,
NYTimes800k, that is 70\% larger than GoodNews~\cite{Biten2019GoodNews} and has
higher-quality articles along with additional image location information. We
observe a performance gain of $6.8\times$ in BLEU-4 ($0.89 \rightarrow 6.05$)
and $4.1\times$ in CIDEr ($13.1 \rightarrow 53.8$) compared to previous
work~\cite{Biten2019GoodNews}. On both datasets we observe consistent gains for
each new component in our language, vision, and knowledge-aware system. We also
find that our model generates names not seen during training, resulting in
linguistically richer captions, which are closer in length (mean 15 words) to
the ground truth (mean 18 words) than the previous state of the art (mean 10
words).




Our main contributions include:

\begin{enumerate}
	\itemsep0em
   \item A new captioning model that incorporates transformers, an
   attention-centric language model, byte-pair encoding, and attention over
   four different modalities (text, images, faces, and objects). \eat{We show
   that our model achieves state-of-the-art results with a significant margin
   over previous methods, and }

   \item Significant performance gains over all metrics, with associated
   ablation studies quantifying the contributions of our main modeling
   components using BLEU-4, CIDEr, precision \& recall of named entities and
   rare proper nouns, and linguistic quality metrics.

	\item NYTimes800k, the largest news image captioning dataset to date,
	containing 445K articles and 793K images with captions from The New York
	Times spanning 14 years. NYTimes800k builds and improves upon the recently
	proposed GoodNews dataset~\cite{Biten2019GoodNews}. It has 70\% more
	articles and includes image locations within the article text. The dataset,
	code, and pre-trained models are available on
	GitHub\footnote{\href{https://github.com/alasdairtran/transform-and-tell}{https://github.com/alasdairtran/transform-and-tell}}.
\end{enumerate}


\section{Related Works}

A popular design choice for image captioning systems involves using a
convolutional neural network (CNN) as the image encoder and a recurrent neural
network (RNN) with a closed vocabulary as a decoder~\cite{Karpathy2015DeepVA,
   Donahue2015LongTR, Vinyals2015ShowAT}. Attention over image patches using a
multilayer perception was introduced in ``Show, Attend and Tell"
\cite{Xu2015ShowAA}. Further extensions include having the option to not attend
to any image region~\cite{Lu2017KnowingWT} using a bottom-up approach to
propose a region to attend to~\cite{Anderson2017BottomUpAT}, and attending
specifically to object regions~\cite{Wang2019Hierarchical} and visual concepts
\cite{You2016ImageCW,Li2019Boosted,Wang2019Hierarchical} identified in the
image.



News image captioning includes the article text as input and focuses on the
types of images used in news articles. A key challenge here is to generate
correct entity names, especially rare ones. Existing approaches include
extractive methods that use n-gram models to combine existing phrases
\cite{Feng2013AutomaticCG} or simply retrieving the most representative
sentence~\cite{Tariq2017ACE} in the article. Ramisa
\etal~\cite{Ramisa2016BreakingNewsAA} built an end-to-end LSTM decoder that
takes both the article and image as inputs, but the model was still unable to
produce names that were not seen during training.

To overcome the limitation of a fixed-size vocabulary, template-based methods
have been proposed. An LSTM first generates a template
sentence with placeholders for named entities, e.g. ``PERSON speaks at BUILDING
in DATE.''~\cite{Biten2019GoodNews}. Afterwards the best candidate for each
placeholder is chosen via a knowledge graph of entity
combinations~\cite{Lu2018EntityAI}, or via sentence
similarity~\cite{Biten2019GoodNews}. One key difference between our proposed
model and previous approaches~\cite{Biten2019GoodNews,Lu2018EntityAI}
is that our model can generate a caption with named entities directly without
using an intermediate template.

\begin{figure*}[t]
   \begin{center}
      \includegraphics[width=\linewidth]{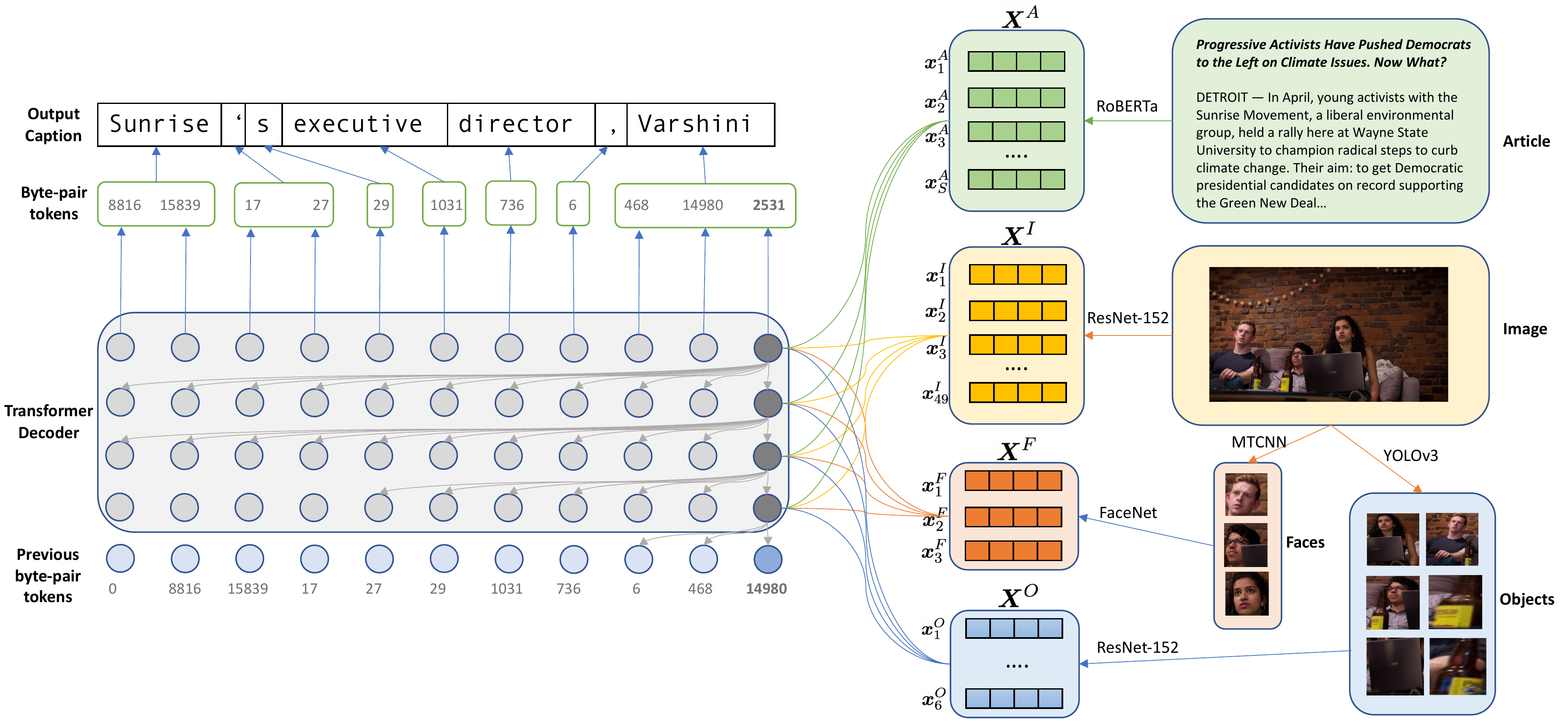}
   \end{center}

   \capmoveup
   \caption{Overview of the Transform and Tell model. Left: Decoder with four
      transformer blocks; Right: Encoder for article, image, faces, and
      objects. The decoder takes embeddings of byte-pair tokens as input (blue
      circles at the bottom). For example, the input in the final time step,
      14980, represents ``arsh'' in ``Varshini'' from the previous time step.
      The grey arrows show the convolutions in the final time step in each
      block. Colored arrows show attention to the four domains on the right:
      article text (green lines), image patches (yellow lines), faces (orange
      lines), and objects (blue lines). The final decoder outputs are byte-pair
      tokens, which are then combined to form whole words and punctuations.}
   \postfigmoveup
   \label{fig:model}
\end{figure*}

One tool that has seen recent successes in many natural language processing
tasks are transformer networks. Transformers have been shown to consistently
outperform RNNs in language modeling~\cite{Radford2019LanguageMA},
story generation~\cite{Fan2018HierarchicalNS},
summarization~\cite{Subramanian2019OnEA}, and machine
translation~\cite{Bojar2018Findings}. In particular, transformer-based models
such as BERT~\cite{Devlin2019BERT}, XLM~\cite{Lample2019CrosslingualLM},
XLNet~\cite{Yang2019XLNetGA}, RoBERTa~\cite{Liu2019RoBERTaAR}, and
ALBERT~\cite{Lan2019ALBERT} are able to produce high level text representations
suitable for transfer learning. Furthermore, using byte-pair encoding
(BPE)~\cite{Sennrich2015NeuralMT} to represent uncommon words as a sequence of
subword units enables transformers to function in an open vocabulary setting.
To date the only image captioning work that uses BPE
is~\cite{Zhao2019InformativeIC}, but they did not use it for rare named
entities as these were removed during pre-processing. In contrast we explicitly
examine BPE for generating rare names and compare it to template-based methods.

Transformers have been shown to yield competitive results in generating generic
MS COCO captions~\cite{Zhu2018CaptioningTW, Li2019Boosted}. Zhao
\etal~\cite{Zhao2019InformativeIC} have gone further and trained transformers
to produce some named entities in the Conceptual Captions
dataset~\cite{Sharma2018ConceptualCA}. However, the authors used web-entity labels, extracted using
Google Cloud Vision API, as inputs to the model. In our work, we do not
explicitly give the model a list of entities to appear in the caption. Instead
our model automatically identifies relevant entities from the provided news
article.



\section{The Transform and Tell Model}

Our model consists of a set of pretrained encoders and a decoder, as
illustrated in Figure \ref{fig:model}. The encoders
(Section~\ref{ssec:encoder}) generate high-level vector representations of the
images, faces, objects, and article text. The decoder
(Section~\ref{ssec:decoder}) attends over these representations to generate a
caption at the sub-word level.

\subsection{Encoders}
\label{ssec:encoder}

\noindent\textbf{Image Encoder:} An overall image representation is obtained
from a ResNet-152~\cite{He2016ResNet} model pre-trained on ImageNet. We use the
output of the final block before the pooling layer as the image representation.
This is a set of 49 different vectors $\bx^I_{i} \in \mathbb{R}^{2048}$ where
each vector corresponds to a separate image patch after the image is divided
into equally-sized 7 by 7 patches. This gives us the set $\bX^I = \{\bx^I_{i}
\in \mathbb{R}^{D^I}\}_{i=1}^{M^I}$, where $D^I = 2048$ and $M^I = 49$ for
ResNet-152. Using this representation allows the decoder to attend to different
regions of the image, which is known to improve performance in other image
captioning tasks~\cite{Xu2015ShowAA} and has been widely adopted.

\noindent\textbf{Face Encoder:} We use MTCNN~\cite{Zhang2016JointFD} to detect
face bounding boxes in the image. We then select up to four faces since the
majority of the captions contain at most four people's names (see
Section~\ref{sec:dataset}). A vector representation of each face is obtained by
passing the bounding boxes to FaceNet~\cite{Schroff2015FaceNetAU}, which was
pre-trained on the VGGFace2 dataset~\cite{Cao2017VGGFace2AD}. The resulting set
of face vectors for each image is
$\bX^{F} = \{\bx^F_{i} \in \mathbb{R}^{D^F}\}_{i=1}^{M^F}$, where $D^F = 512$ for FaceNet and $M^F$ is the
number of faces. If there are no faces in the image, $\bX^{F}$ is an empty set.

Even though the faces are extracted from the image, it is useful to consider
them as a separate input domain. This is because a specialized face embedding
model is tuned for identifying people and thus can help the
decoder to generate more accurate named entities.


\eat{We then select the top $M$ faces and feed them through a
FaceNet model~\cite{Schroff2015FaceNetAU}, pretrained on the VGGFace2 dataset
\cite{Cao2017VGGFace2AD}, to obtain a face embedding $\bx_{F} =
   \{\bx_{F,i} \in \mathbb{R}^{512}\}_{i=0}^M$.}

\noindent\textbf{Object Encoder:} We use YOLOv3~\cite{Redmon2018YOLOv3AI} to
detect object bounding boxes in the image. We filter out objects with a
confidence less than 0.3 and select up to 64 objects with the highest
confidence scores to feed through a ResNet-152 pretrained on ImageNet. In
contrast to the image encoder, we take the output after the pooling layer as
the representation for each object. This gives us a set of object vectors
$\bX^{O} = \{\bx^O_{i} \in \mathbb{R}^{D^O}\}_{i=1}^{M^O}$, where $D^O = 2048$
for ResNet-152 and $M^O$ is the number of objects.

\noindent\textbf{Article Encoder:} To encode the article text we use
RoBERTa~\cite{Liu2019RoBERTaAR}, a recent improvement over the popular
BERT~\cite{Devlin2019BERT} model. RoBERTa is a pretrained language
representation model providing contextual embeddings for text. It consists of
24 layers of bidirectional transformer blocks.

Unlike GloVe~\cite{Pennington2014Glove} and word2vec
\cite{Mikolov2013DistributedRO} embeddings, where each word has exactly one
representation, the bidirectionality and the attention mechanism in the
transformer allow a word to have different vector representations depending on
the surrounding context.

The largest GloVe model has a vocabulary size of 1.2 million. Although this is
large, many rare names will still get mapped to the unknown token. In contrast,
RoBERTa uses BPE~\cite{Sennrich2015NeuralMT,Radford2019LanguageMA} which can
encode any word made from Unicode characters. In BPE, each word is first broken
down into a sequence of bytes. Common byte sequences are then merged using a
greedy algorithm. Following \cite{Radford2019LanguageMA},
our vocabulary consists of 50K most common byte sequences.

Inspired by Tenney \etal~\cite{Tenney2019BertRT} who showed that different
layers in BERT represent different steps in the traditional NLP pipeline, we
mix the RoBERTa layers to obtain a richer representation. Given an input of
length $M^T$, the pretrained RoBERTa encoder will return 25 sequences of
embeddings, $\bG = \{\bg_{\ell i} \in \mathbb{R}^{2048} : \ell \in \{ 0, 1,...,
24 \}, i \in \{1, 2,..., M^T\}\}$. This includes the initial uncontextualized
embeddings and the output of each of the 24 transformer layers. We take a
weighted sum across all layers to obtain the article embedding $\bx^A_{i}$:
\begin{IEEEeqnarray}{lCl}
   \bx^A_{i} &=& \sum_{\ell=0}^{24} \alpha_\ell \, \bg_{\ell i}
\end{IEEEeqnarray}
where $\alpha_\ell$ are learnable weights.

Thus our RoBERTa encoder produces the set of token embeddings $\bX^{A} =
\{\bx^A_{i} \in \mathbb{R}^{D^T}\}_{i=1}^{M^T}$, where $D^T = 1024$ in RoBERTa.

\subsection{Decoder}
\label{ssec:decoder}

The decoder is a function that generates caption tokens sequentially. At time
step $t$, it takes as input: the embedding of the token generated in the
previous step, $\bz_{0t} \in \mathbb{R}^{D^E}$ where $D^E$ is the hidden size;
embeddings of all other previously generated tokens $\bZ_{0<t} = \{\bz_{00},
\bz_{01}, ..., \bz_{0t-1} \}$; and the context embeddings $\bX^I$, $\bX^A$,
$\bX^F$, and $\bX^O$ from the encoders. These inputs are then fed through $L$
transformer blocks:
\begin{IEEEeqnarray}{lCl}
   \bz_{1t} &=& \text{Block}_1 (\bz_{0t} | \bZ_{0<t}, \bX^I, \bX^A, \bX^F, \bX^O) \\
   \bz_{2t} &=& \text{Block}_2 (\bz_{1t} | \bZ_{1<t}, \bX^I, \bX^A, \bX^F, \bX^O) \\
   &\dots& \notag \\
   \bz_{Lt} &=& \text{Block}_L (\bz_{L-1t} | \bZ_{L-1<t}, \bX^I, \bX^A, \bX^F, \bX^O)
\end{IEEEeqnarray}
where $\bz_{\ell t}$ is the output of the $\ell$\textsuperscript{th}
transformer block at time step $t$. The final block's output $\bz_{Lt}$ is used
to estimate $p(y_t)$, the probability of generating the $t$th token in the
vocabulary via adaptive softmax~\cite{Grave2016EfficientSA}:
\begin{IEEEeqnarray}{lCl}
   p(y_t) &=& \text{AdaptiveSoftmax}(\bz_{Lt})
\end{IEEEeqnarray}
By dividing the vocabulary into three clusters based on frequency---5K, 15K,
and 30K---adaptive softmax makes training more efficient since most of the
time, the decoder only needs to compute the softmax over the first cluster
containing the 5,000 most common tokens.

In the following two subsections, we will describe the transformer block in
detail. In each block, the conditioning on past tokens is achieved using
dynamic convolutions, and the conditioning on the contexts
is achieved using multi-head attention.


\noindent\textbf{Dynamic Convolutions:}
Introduced by Wu \etal~\cite{Wu2018PayLA}, the goal of dynamic convolution is
to provide a more efficient alternative to
self-attention~\cite{Vaswani2017AttentionIA} when attending to past tokens. At
block $\ell + 1$ and time step $t$, we have the input $\bz_{\ell t} \in
\mathbb{R}^{D^E}$. Given kernel size $K$ and
$H$ attention heads, for each head $h \in \{1, 2, ..., H\}$, we first project
the current and last $K-1$ steps using a feedforward layer to obtain $\bz_{\ell
h j}' \in \mathbb{R}^{D^E/H}$:
\begin{IEEEeqnarray}{lCl}
   \bz_{\ell h j}' &=& \text{GLU}(\bW^Z_{\ell h} \, \bz_{\ell j} + \bb^Z_{\ell h})
\end{IEEEeqnarray}
for $j \in \{t - K + 1,t - K + 2,...,t\}$. Here GLU is the gated linear unit
activation function~\cite{Dauphin2017GLU}. The output of each head's dynamic
convolution is the weighted sum of these projected values:
\begin{IEEEeqnarray}{lCl}
   \tz_{\ell h t} &=& \sum_{j=t - K + 1}^{t} \gamma_{\ell hj} \, \bz_{\ell h j}'
\end{IEEEeqnarray}
where the weight $\gamma_{\ell hj}$ is a linear projection of the input (hence
the term ``dynamic"),
followed by a softmax over the kernel window:
\begin{IEEEeqnarray}{lCl}
   \gamma_{\ell hj} &=& \text{Softmax} \left( (\bw^\gamma_{ \ell h})^T \,
   \bz'_{\ell h j} \right)
\end{IEEEeqnarray}
The overall output is the concatenation of all the head outputs, followed by a
feedforward with a residual connection and layer
normalization~\cite{Ba2016LayerN}, which does a z-score normalization across
the feature dimension (instead of the batch dimension as in batch
normalization~\cite{Ioffe2015BatchNorm}):
\begin{IEEEeqnarray}{lCl}
   \tz_{\ell t} &=& [ \tz_{\ell 1 t}, \tz_{\ell 2 t},..., \tz_{\ell H t} ] \\
   \bd_{\ell t} &=& \text{LayerNorm}\left( \bz_{\ell t} +
   \bW^{\tz}_{\ell} \, \tz_{\ell t} + \bb^{\tz}_{\ell} \right)
\end{IEEEeqnarray}
The output $\bd_{\ell t}$ can now be used to attend over the context embeddings.


\noindent\textbf{Multi-Head Attention:}
The multi-head attention mechanism~\cite{Vaswani2017AttentionIA} has been the
standard method to attend over encoder outputs in transformers. In our setting,
we need to attend over four context domains---images, text, faces, and objects.
As an example, we will go over the image attention module, which consists
of $H$ heads. Each head $h$ first does a linear projection of $\bd_{\ell t}$
and the image embeddings $\bX^I$ into a query $\bq^I_{\ell h t} \in
\mathbb{R}^{D^E/H}$, a set of keys $\bK^I_{\ell h t} = \{\bk^I_{\ell h t i}
\in \mathbb{R}^{D^E/H} \}_{i=1}^{M^I}$, and the corresponding values
$\bV^I_{\ell h t} = \{\bv^{I}_{\ell h t i} \in \mathbb{R}^{D^E/H}
\}_{i=1}^{M^I}$:
\begin{IEEEeqnarray}{lCl}
   \bq^{I}_{\ell ht} &=& \bW^{IQ}_{ \ell h} \, \bd_{\ell t} \\
   \bk^{I}_{\ell h i} &=& \bW^{IK}_{ \ell h} \, \bx^{I}_{ i}
   \qquad \forall i \in \{1, 2, ..., M^I\}\\
   \bv^{I}_{ \ell h i} &=& \bW^{IV}_{ \ell h} \, \bx^{I}_{ i}
   \qquad \forall i \in \{1, 2, ..., M^I\}
\end{IEEEeqnarray}
Then the attended image for each head is the weighted sum of the values, where
the weights are obtained from the dot product between the query and key:
\begin{IEEEeqnarray}{lCl}
   \lambda^{I}_{ \ell h t i} &=&\text{Softmax}\left(\bK^{I}_{\ell h} \, \bq^{I}_{ \ell ht} \right)_i\\
   \bx^{'I}_{ \ell h t} &=& \sum_{i = 1}^{M^I}
   \lambda^{I}_{ \ell h t i} \, \bv^{I}_{ \ell h t i}
\end{IEEEeqnarray}
The attention from each head is then concatenated into $\bx^{'I}_{\ell t} \in
   \mathbb{R}^{D^E}$:
\begin{IEEEeqnarray}{lCl}
   \bx^{'I}_{\ell t} &=& [\tx^{I}_{\ell 1 t}, \tx^{I}_{\ell 2 t}, ...,
      \tx^{I}_{\ell H t}]
\end{IEEEeqnarray}
and the overall image attention $\tx^{I}_{\ell t} \in \mathbb{R}^{D^E}$ is obtained
after adding a residual connection and layer normalization:
\begin{IEEEeqnarray}{lCl}
   \tx^{I}_{ \ell t} &=& \text{LayerNorm}(\bd_{\ell t} + \bx^{'I}_{\ell t})
\end{IEEEeqnarray}
We use the same multi-head attention mechanism (with different weight matrices)
to obtain the attended article $\tx^A_{ \ell t}$, faces $\tx^F_{\ell t}$, and
objects $\tx^O_{\ell t}$. These four are finally concatenated and fed through a
feedforward layer:
\begin{IEEEeqnarray}{lCl}
   \tx^C_{\ell t} &=& [\tx^{I}_{ \ell t}, \tx^A_{\ell t}, \tx^F_{\ell t}, \tx^O_{\ell t}] \\
   \tx^{C'}_{\ell t} &=& \bW^C_{\ell} \, \tx^C_{\ell t} + \bb^C_{\ell} \\
   \tx^{C''}_{\ell t} &=& \text{ReLU}(\bW^{C'}_{\ell} \, \tx^{C'}_{\ell t} + \bb^{C'}_{\ell} )\\
   \bz_{\ell + 1\,t} &=& \text{LayerNorm}(\tx^{C'}_{\ell t} + \bW^{C''}_{\ell} \,
   \tx^{C''}_{\ell t} + \bb^{C''}_{\ell})
\end{IEEEeqnarray}
The final output $\bz_{\ell + 1\,t} \in \mathbb{R}^{D^E}$ is used as the input
to the next transformer block.

\section{News Image Captioning Datasets}
\label{sec:dataset}
We describe two datasets that contain news articles, images, and captions. The
first dataset, GoodNews, was recently proposed in Biten
\etal~\cite{Biten2019GoodNews}, while the
second dataset, NYTimes800k, is our contribution.

\noindent{\bf GoodNews:}
The GoodNews dataset was previously the largest dataset for news image
captioning~\cite{Biten2019GoodNews}. Each example in the dataset is a triplet
containing an article, an image, and a caption. Since only the article text,
captions, and image URLs are publicly released, the images need to be
downloaded from the original source. Out of the 466K image URLs provided by
\cite{Biten2019GoodNews}, we were able to download 463K images, or 99.2\% of
the original dataset---the remaining are broken links.

We use this 99.2\% sample of GoodNews and the train-validation-test split
provided by~\cite{Biten2019GoodNews}. There are 421K training, 18K validation,
and 23K test captions. Note that this split was performed at the level of
captions, so it is possible for a training and test caption to share the same
article text (since articles have multiple images).

We observe several issues with GoodNews that may limit a system's
ability to generate high-quality captions. Many of the articles in GoodNews are
partially extracted because the generic article extraction library
failed to recognize some of the HTML tags specific to The New York Times.
Importantly, the missing text often included the first few paragraphs which
frequently contain important information for captioning images. In addition
GoodNews contains some non-English articles and captioned images from the
recommendation sidebar which are not related to the main article.

\begin{table}[t]
	\caption {Summary of news captioning datasets}
	\label{tab:datasets}
	\centering
	\begin{tabularx}{\linewidth}{lXX}
		\toprule
		  & GoodNews  &   NYTimes800k \\
		\midrule
      Number of articles & 257 033 & 444 914 \\
      Number of images   & 462 642 & 792 971 \\
      Average article length & 451 & 974 \\
      Average caption length & 18 & 18 \\
      Collection start month & Jan 10 & Mar 05\\
      Collection end month & Mar 18 & Aug 19 \\
      \midrule
      \multicolumn{2}{l}{\% of caption words that are}  \\
      \quad -- nouns & 16\% & 16\% \\
      \quad -- pronouns & 1\% & 1\% \\
      \quad -- proper nouns & 23\% & 22\% \\
      \quad -- verbs & 9\% & 9\%  \\
      \quad -- adjectives & 4\% & 4\% \\
      \quad -- named entities & 27\% & 26\% \\
      \quad\quad -- people's names & 9\% & 9\% \\
      \midrule
      \% of captions with \\
      \quad -- named entities & 97\% & 96\% \\
      \quad\quad -- people's names & 68\% & 68\% \\
		\bottomrule
	\end{tabularx}
\end{table}

\noindent{\bf NYTimes800k:}
The aforementioned issues motivated us to construct NYTimes800k, a 70\% larger
and more complete dataset of New York Times articles, images, and captions. We
used The New York Times public
API\footnote{\href{https://developer.nytimes.com/apis}{https://developer.nytimes.com/apis}}
for the data collection and developed a custom parser to resolve the missing
text issue in GoodNews. The average article in NYTimes800k is 963 words long,
whereas the average article in GoodNews is 451 words long. Our parser also
ensures that NYTimes800k only contains English articles and images that are
part of the main article. Finally, we also collect information about where an
image is located in the corresponding article. Most news articles have one
image at the top that relates to the key topic. However 39\% of the articles
have at least one more image somewhere in the middle of text. The image
placement and the text surrounding the image is important information for
captioning as we will show in our evaluations. Table~\ref{tab:datasets}
presents a comparison between GoodNews and NYTimes800k.

\begin{figure}[t]
   \begin{center}
   \includegraphics[width=0.99\linewidth]{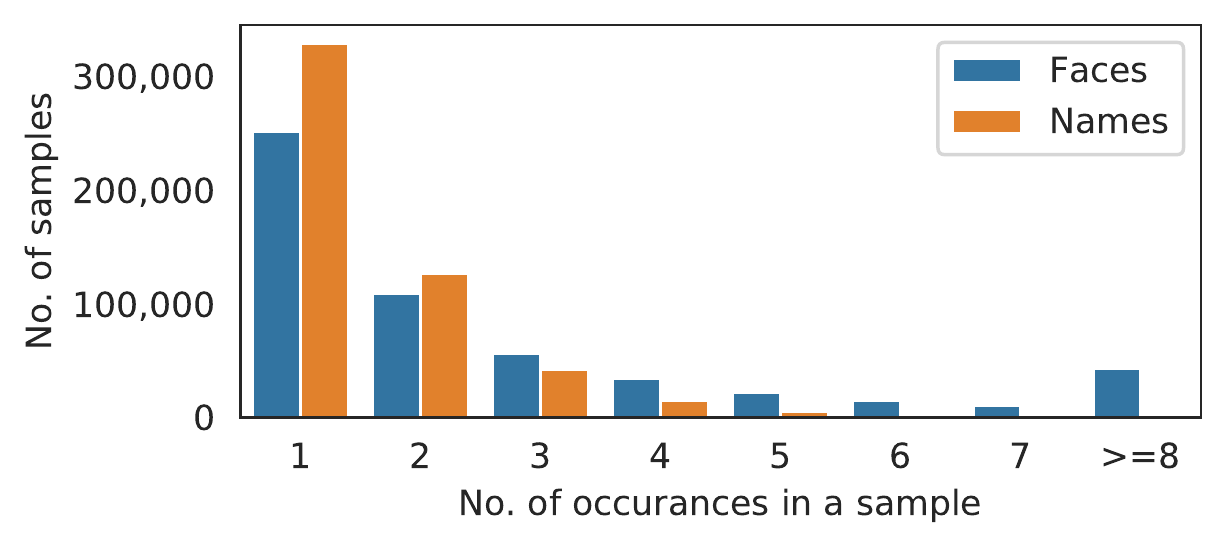}
   \end{center}
  \capmoveup
      \caption{Co-occurrence of faces and people's names in NYTimes800k
               training data. The blue bars count how many images containing a
               certain number of faces. The orange bars count how many captions
               containing a certain number of people's names.}
   \vspace{-3.5mm}
   \label{fig:faces}
\end{figure}

Entities play an important role in NYTimes800k, with 97\% of captions
containing at least one named entity. The most popular entity type are names of
people, comprising a third of all named entities (see the supplementary
material for a detailed breakdown of entity types). Furthermore, 71\% of
training images contain at least one face and 68\% of training captions mention
at least one person's name. Figure \ref{fig:faces} provides a further breakdown
of the co-occurrence of faces and people's names. One important observation is
that 99\% of captions contain at most four names.

We split the training, validation, and test sets according to time, as shown in
Table~\ref{tab:splits}. Compared to the random split used in GoodNews,
splitting by time allows us to study the model performance on novel news events
and new names, which might be important in a deployment scenario. Out of the
100K proper nouns in our test captions, 4\% never appear in any training
captions.


\secmoveup
\section{Experiments}
This section describes settings for neural network learning, baselines and
evaluation metrics, followed by a discussion of key results.

\secmoveup
\subsection{Training Details}
\label{ssec:training_details}



Following Wu \etal~\cite{Wu2018PayLA}, we set the hidden size $D^E$ to 1024;
the number of heads $H$ to 16; and the number of transformer blocks $L$ to four
with kernel sizes 3, 7, 15, and 31, respectively. For parameter optimization we
use the adaptive gradient algorithm Adam~\cite{Kingma2015Adam} with the
following parameter: $\beta_1 = 0.9, \beta_2 = 0.98, \epsilon = 10^{-6}$. We
warm up the learning rate in the first 5\% of the training steps to $10^{-4}$,
and decay it linearly afterwards. We apply $L_2$ regularization to all network
weights with a weight decay of $10^{-5}$ and using the
fix~\cite{Loshchilov2018DecoupledWD} that decouples the learning rate from the
regularization parameter. We clip the gradient norm at 0.1. We use a maximum
batch size of 16 and training is stopped after the model has seen 6.6 million
examples. This is equivalent to 16 epochs on GoodNews and 9 epochs on
NYTimes800k.

The training pipeline is written in PyTorch~\cite{Paszke2017Automatic} using
the AllenNLP framework~\cite{Gardner2017AllenNLP}. The RoBERTa model and
dynamic convolution code are adapted from fairseq~\cite{Ott2019Fairseq}.
Training is done with mixed precision to reduce the memory footprint and allow
our full model to be trained on a single GPU. The full model takes 5 days to
train on one Titan V GPU and has 200 million trainable parameters---see the
supplementary material for the size of each model variant.

\secmoveup
\subsection{Evaluation Metrics}

We use BLEU-4~\cite{Papineni2002Bleu} and CIDEr~\cite{Vedantam2015CIDEr} scores
as they are standard for evaluating image captions. These are obtained using
the COCO caption evaluation
toolkit\footnote{\href{https://github.com/tylin/coco-caption}
{https://github.com/tylin/coco-caption}}. The supplementary material
additionally reports BLEU-1, BLEU-2, BLEU-3, ROUGE \cite{Lin2004ROUGE}, and
METEOR~\cite{Denkowski2014Meteor}. Note that CIDEr is particularly suited for
evaluating news captioning models as it puts more weight than other metrics on
uncommon words. In addition, we evaluate the precision and recall on named
entities, people's names, and rare proper names. Named entities are identified
in both the ground-truth captions and the generated captions using SpaCy. We
then count exact string matches between the ground truths and generated
entities. For people's names we restrict the set of named entities to those
marked as PERSON by the SpaCy parser. Rare proper nouns are nouns that appear
in a test caption but not in any training caption.

\begin{table}[t]
	\caption {NYTimes800k training, validation, and test splits}
	\label{tab:splits}
	\centering
	\begin{tabularx}{\linewidth}{lXXX}
		\toprule
		  & Training  &   Validation & Test \\
		\midrule
      Number of articles & 433 561 & 2 978 & 8 375 \\
      Number of images  & 763 217 & 7 777 & 21 977 \\
      Start month & Mar 15 & May 19 & Jun 19 \\
      End month & Apr 19 & May 19 & Aug 19 \\
		\bottomrule
	\end{tabularx}
\end{table}

\begin{table*}[t]

	\caption {Results on GoodNews (rows 1--10) and NYTimes800k (rows 11--19).
		We report BLEU-4, ROUGE, CIDEr, and precision (P) \& recall (R)  of
		named entities, people's names, and rare proper nouns. Precision and
		recall are expressed as percentages. Rows 1--2 contain previous
		state-of-the-art results \cite{Biten2019GoodNews}. Rows 3--5 and 11--13
		are ablation studies where we swap the Transformer with an LSTM and/or
		RoBERTa with GloVe. These models only have the image attention (IA).
		Rows 6 \& 14 are our baseline RoBERTa transformer language model that
		only has the article text (and not the image) as inputs. Building on
		top of this, we first add attention over image patches (rows 7 \& 15).
		We then take a weighted sum of the RoBERTa embeddings (rows 8 \& 16)
		and attend to the text surrounding the image instead of the first 512
		tokens of the article (row 17). Finally we add attention over faces
		(rows 9 \& 18) and objects (rows 10 \& 19) in the image.}

	\label{tab:results}
	\centering
	\begin{tabularx}{\textwidth}{llXXX XX XX XX}
		\toprule
		 &
		 & \multirow{2}{*}{\mbox{\small{BLEU-4}}}
		 & \multirow{2}{*}{\small{ROUGE}}
		 & \multirow{2}{*}{\small{CIDEr}}
		 & \multicolumn{2}{l}{\small{Named entities}}
		 & \multicolumn{2}{l}{\small{People's names}}
		 & \multicolumn{2}{l}{\small{Rare proper nouns}}                                                                                                                                                     \\
		 &                                                   &               &               &               & \small{P}     & \small{R}     & \small{P}     & \small{R}     & \small{P}     & \small{R}     \\
		\midrule
		\multirow{9}{*}{\rotatebox[origin=c]{90}{GoodNews}}
		 & (1) Biten (Avg + CtxIns)~\cite{Biten2019GoodNews} & 0.89          & 12.2          & 13.1          & 8.23          & 6.06          & 9.38          & 6.55          & 1.06          & 12.5          \\
		 & (2) Biten (TBB + AttIns)~\cite{Biten2019GoodNews} & 0.76          & 12.2          & 12.7          & 8.87          & 5.64          & 11.9          & 6.98          & 1.58          & 12.6          \\
		\cmidrule{2-11}

		 & (3) LSTM + GloVe + IA                             & 1.97          & 13.6          & 13.9          & 10.7          & 7.09          & 9.07          & 5.36          & 0             & 0             \\
		 & (4) Transformer + GloVe + IA                      & 3.48          & 17.0          & 25.2          & 14.3          & 11.1          & 14.5          & 10.5          & 0             & 0             \\
		 & (5) LSTM + RoBERTa + IA                           & 3.45          & 17.0          & 28.6          & 15.5          & 12.0          & 16.4          & 12.4          & 2.75          & 8.64          \\
		\cmidrule{2-11}

		 & (6) Transformer + RoBERTa                         & 4.60          & 18.6          & 40.9          & 19.3          & 16.1          & 24.4          & 18.7          & 10.7          & 18.7          \\
		 & (7) \quad + image attention                       & 5.45          & 20.7          & 48.5          & 21.1          & 17.4          & 26.9          & 20.7          & 12.2          & 20.9          \\
		 & (8) \quad\quad + weighted RoBERTa                 & 6.0           & 21.2          & 53.1          & 21.8          & 18.5          & 28.8          & 22.8          & \textbf{16.2} & 26.0          \\
		 & (9) \quad\quad\quad + face attention              & \textbf{6.05} & \textbf{21.4} & \textbf{54.3} & 22.0          & 18.6          & \textbf{29.3} & \textbf{23.3} & 15.5          & 24.5          \\
		 & (10) \quad\quad\quad\quad + object attention      & \textbf{6.05} & \textbf{21.4} & 53.8          & \textbf{22.2} & \textbf{18.7} & 29.2          & 23.1          & 15.6 & \textbf{26.3} \\

		\midrule
		\midrule
		\multirow{8}{*}{\rotatebox[origin=c]{90}{NYTimes800k}}
		 & (11) LSTM + GloVe + IA                            & 1.77          & 13.1          & 12.1          & 10.2          & 7.24          & 8.83          & 5.73          & 0             & 0             \\
		 & (12) Transformer + GloVe + IA                     & 2.75          & 15.9          & 20.3          & 13.2          & 10.8          & 13.2          & 9.66          & 0             & 0             \\
		 & (13) LSTM + RoBERTa + IA                          & 3.29          & 16.1          & 24.9          & 15.1          & 12.9          & 17.7          & 14.4          & 7.47          & 9.50          \\
		\cmidrule{2-11}
		 & (14) Transformer + RoBERTa                        & 4.26          & 17.3          & 33.9          & 17.8          & 16.3          & 23.6          & 19.7          & 21.1          & 16.7          \\
		 & (15) \quad + image attention                      & 5.01          & 19.4          & 40.3          & 20.0          & 18.1          & 28.2          & 23.0          & 24.3          & 19.3          \\
		 & (16) \quad\quad + weighted RoBERTa                & 5.75          & 19.9          & 45.1          & 21.1          & 19.6          & 29.7          & 25.4          & 29.6          & 22.8          \\
		 & (17) \quad\quad\quad + location-aware             & 6.36          & 21.4          & 52.8          & 24.0          & 21.9          & 35.4          & 30.2          & 33.8          & \textbf{27.2} \\
		 & (18) \quad\quad\quad\quad + face attention        & 6.26          & 21.5          & 53.9          & 24.2          & 22.1          & 36.5          & 30.8          & 33.4          & 26.4          \\
		 & (19) \quad\quad\quad\quad\quad + object attention & \textbf{6.30} & \textbf{21.7} & \textbf{54.4} & \textbf{24.6} & \textbf{22.2} & \textbf{37.3} & \textbf{31.1} & \textbf{34.2} & 27.0          \\

		\bottomrule
	\end{tabularx}
\end{table*}

\subsection{Baselines and Model Variants}

We show two previous state-of-the-art models: \textit{Biten (Avg + CtxIns)} and
\textit{Biten (TBB + AttIns)}~\cite{Biten2019GoodNews}. To provide a fair
comparison we used the full caption results released by Biten
\etal~\cite{Biten2019GoodNews} and re-evaluated with our evaluation pipeline on
a slightly smaller test set (a few test images are no longer available due to
broken URLs). The final metrics are the same as originally reported if rounded
to the nearest whole number.

We evaluate a few key modeling choices: the decoder type (\textit{LSTM} vs
\textit{Transformer}), the text encoder type (\textit{GloVe} vs
\textit{RoBERTa} vs \textit{weighted RoBERTa}), and the additional context
domains (\textit{location-aware}, \textit{face attention}, and \textit{object
attention}). The \textit{location-aware} models select the 512 tokens
surrounding the image instead of the first 512 tokens of the article. Note that
all our models use BPE in the decoder with adaptive softmax. We ensure that the
total number of trainable parameters for each model is within 7\% of one
another (148 million to 159 million), with the exception of \textit{face
attention} (171 million) and \textit{object attention} (200 million) since the
latter two have extra multi-head attention modules. The results reported over
GoodNews are based on a model trained solely on GoodNews, using the original
random split of \cite{Biten2019GoodNews} for easier comparison to previous work.

\eat{The due primarily to broken image links the numbers in the table are
	slightly different to what was previously
	reported~\cite{Biten2019GoodNews}.
	Note that
	the numbers reported here are slightly different from the original paper
	since
	we had to remove a few samples from the test set where the image is no
	longer
	available.~\cite{Biten2019GoodNews} also did some post-processing on the
	ground-truth captions such as removing contractions and non-ASCII
	characters,
	both of which we did not do. Despite these differences, the final metrics
	are the same if rounded to the nearest whole number.}

\secmoveup
\subsection{Results and Discussion}

\eat{There is a strong correlation between of all these metrics, and in
	general, we
	mainly look at CIDEr since it uses Term Frequency Inverse Document Frequency
	(TF-IDF) to put more importance on less common words such as entity names.
	Table~\ref{tab:names} shows the recall and precision of the named entities,
	people's names, and rare proper nouns.}

\begin{figure*}[t]
	\begin{center}
		\includegraphics[width=\linewidth]{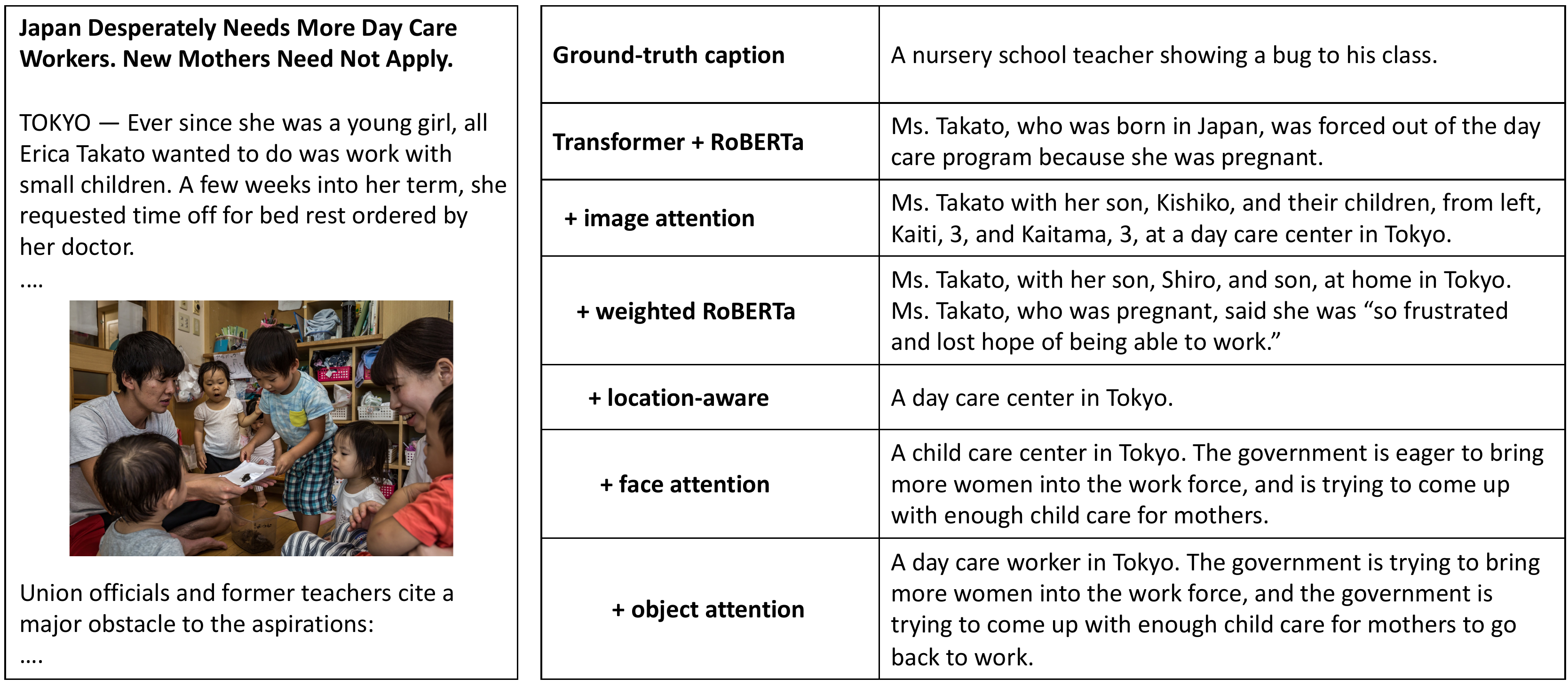}
	\end{center}
	\capmoveup
	\caption{An example article (left) and the corresponding news captions
		(right) from the NYTimes800k test set. The model with no access to the
		image makes a sensible but incorrect guess that the image is about Ms.
		Takato. Since the image appears in the middle of the article, only the
		location-aware models correctly state that the focus of the image is on
		a day care center.}
	\label{fig:example}
	\smallfigmoveup
\end{figure*}

Table~\ref{tab:results} summarizes evaluation metrics on GoodNews and
NYTimes800k, while Figure~\ref{fig:example} compares generated captions from
different model variants. Our full model (row 10) performs substantially better
than the existing state of the art~\cite{Biten2019GoodNews} across all
evaluation metrics. On GoodNews, the full model yields a CIDEr score of 53.8,
whereas the previous state of the art ~\cite{Biten2019GoodNews} achieved a
CIDEr score of only 13.1.

Our most basic LSTM model (row 3) differs from
Biten~\etal~\cite{Biten2019GoodNews} in that we use BPE in the caption decoder
instead of template generation and filling. The slight improvement in CIDEr
(from 13.1 to 13.9) shows that BPE offers a competitive end-to-end alternative
to the template filling method. This justifies the use of BPE in the remaining
experiments.

Models that encode articles using GloVe embeddings (rows 3--4 and 11--12) are
unable to generate rare proper nouns, giving a precision and recall of 0. This
is because the encoder skips words that are not part of the fixed GloVe
vocabulary. This motivates the switch from GloVe to RoBERTa, which has an
unbounded vocabulary. This switch shows a clear advantage in rare proper noun
generation. On NYTimes800k, even the worst performing model that uses RoBERTa
(row 13) achieves a precision of 7.47\%, a recall of 9.50\%, and a CIDEr gap of
12.8 points over the model without RoBERTa (row 11).

Another important modeling choice is the functional form of the caption
decoder. We find that the Transformer architecture provides a substantial
improvement over the LSTM with respect to all evaluation metrics. For example,
when we swap the LSTM with a Transformer (from row 13 to 15), the CIDEr score
on NYTimes800k jumps from 24.9 to 40.3.

\eat{Which is the largest improvement brought by any single modelling choice.}

Adding attention over faces improves both the recall and precision of people's
names. It has no significant effect on other entity types (see the
supplementary material for a detailed breakdown). Importantly, people's names
are the most common entity type in news captions and so we also see an
improvement in CIDEr. Attention over objects also improves performance on most
metrics, especially on NYTimes800k. More broadly, this result suggests that
introducing specialized vision models tuned to the common types of objects such
as organizations (via logos or landmarks) is a promising future direction to
improve the performance on news image captioning.


The location-aware models (rows 17--19) focus the article context using the
image location in the article, information which is only available in our
NYTimes800k dataset. This simple focusing of context offers a big improvement
to CIDEr, from 45.1 (row 16) to 52.8 (row 17). This suggests a strong
correspondence between an image and the closest text that can be easily
exploited to generate better captions.


\eat{\item Our LSTM model with GloVe embeddings outperforms the previous
	state-of-the-art~\cite{Biten2019GoodNews}, indicating that BPE encoding
	offers a viable alternative to template-based methods.}

\eat{ \item Models which encode articles using GloVe embeddings are unable
	to
	generate rare proper nouns (precision and recall of 0).
	This is because the encoder skips words that are not part of the fixed
	GloVe
	vocabulary. Models that use the BPE based article encoder RoBERTa, and thus
	have an unbounded vocabulary show
	a clear advantage in rare proper noun generation, with precision as high as
	38.3\% at a recall of 39.9\%.}


The supplementary material additionally reports three caption quality metrics:
caption length, type-token ratio (TTR)~\cite{Templin1957CertainLS}, and Flesch
reading ease (FRE)~\cite{Flesch1948,Kincaid1975DerivationON}. TTR is the ratio
of the number of unique words to the total number of words in a caption. The
FRE takes into account the number of words and syllables and produces a score
between 0 and 100, where higher means being easier to read. As measured by
FRE, captions generated by our model exhibit a level of language complexity
that is closer to the ground truths. Additionally, captions generated by our
model are 15 words long on average, which is closer to the ground-truths (18
words) than those generated by the previous state of the art (10
words)~\cite{Biten2019GoodNews}.


\secmoveup
\section{Conclusion}
In this paper, we have shown that by using a carefully selected novel
combination of the latest techniques drawn from multiple sub-fields within
machine learning, we are able to set a new SOTA for news image captioning. Our
model can incorporate real-world knowledge about entities across different
modalities and generate text with better linguistic diversity. The key modeling
components are byte-pair encoding that can output any word, contextualized
embeddings for article text, specialized face \& object encoding, and
transformer-based caption generation. This result provides a promising step for
other image description tasks with contextual knowledge, such as web pages,
social media feeds, or medical documents. Promising future directions include
specialized visual models for a broader set of entities like countries and
organizations, extending the image context from the current article to recent
or linked articles, or designing similar techniques for other image and text
domains.

\secmoveup
\section*{Acknowledgement}

This research was supported in part by the Data to Decisions Cooperative
Research Centre whose activities are funded by the Australian Commonwealth
Government’s Cooperative Research Centres Programme. The research was also
supported in part by the Australian Research Council through project number
DP180101985. We thank NVIDIA for providing us with Titan V GPUs through their
GPU Grant Program.


{\small
\bibliographystyle{ieee_fullname}
\bibliography{main}
}

\clearpage


\section{Supplementary Material}

\subsection{Live Demo}

A live demo of our model is available at
\href{https://transform-and-tell.ml}{https://transform-and-tell.ml}. In the
demo, the user is able to provide the URL to a New York Times article. The
server will then scrape the web page, extract the article and image, and
feed them into our model to generate a caption.

\subsection{Entity Distribution}

\begin{figure}[t]
   \begin{center}
      \includegraphics[width=0.99\linewidth]{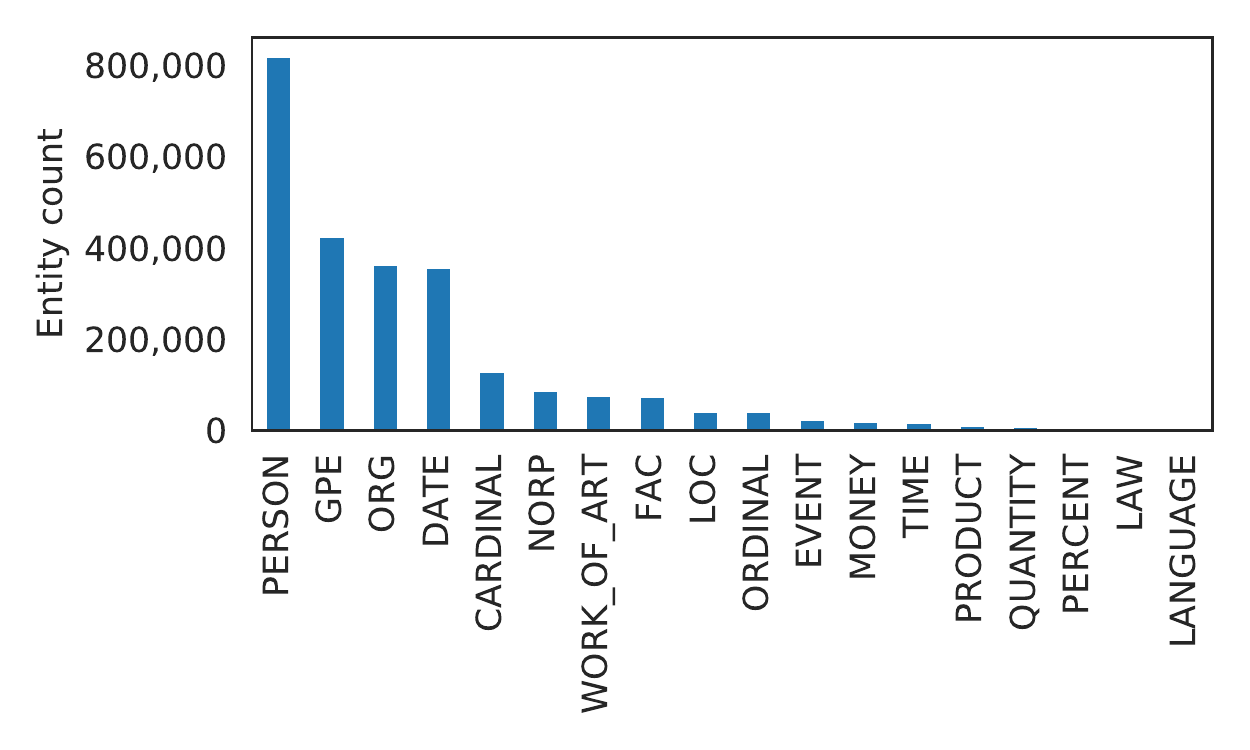}
   \end{center}
   \caption{Entity distribution in NYTimes800k training captions. The four
      most common entity types are people's names, geopolitical
      entities, organizations, and dates.}
   \label{fig:entities}
\end{figure}

Figure \ref{fig:entities} shows how different name entity types are distributed
in the training captions of the NYTimes800k dataset. The four most popular
types are people's names (PERSON), geopolitical entities (GPE), organizations
(ORG), and dates (DATE). Out of these, people's names comprise a third of all
named entities. This motivates us to add a specialized face attention module to
the model.

\subsection{Model Complexity}

\begin{table}[h] \caption {Model complexity. See Table 3 caption in the main
   paper for more explanation of each model variant.}
   \label{tab:models}
   \centering
   \begin{tabularx}{\linewidth}{Xc}
      \toprule
                                                   & No. of Parameters \\
      \midrule
      LSTM + GloVe + IA             & 157M              \\
      Transformer + GloVe + IA        & 148M              \\
      LSTM + RoBERTa + IA     & 159M              \\
      \midrule
      Transformer + RoBERTa                       & 125M              \\
      \quad + image attention (IA)          & 154M              \\
      \quad\quad + weighted RoBERTa                & 154M              \\
      \quad\quad\quad + location-aware             & 154M              \\
      \quad\quad\quad\quad + face attention        & 171M              \\
      \quad\quad\quad\quad\quad + object attention & 200M              \\
      \bottomrule
   \end{tabularx}
\end{table}

Table \ref{tab:models} shows the number of training parameters in each of our
model variants. We ensure that the total number of trainable parameters for
each model is within 7\% of one another (148 million to 159 million), with the
exception of the model with face attention (171 million) and with object
attention (200 million) since the latter two have extra multi-head attention
modules.

\subsection{Further Experimental Results}

Table \ref{tab:results} reports BLEU-1, BLEU-2, BLEU-3,
BLEU-4~\cite{Papineni2002Bleu} ROUGE~\cite{Lin2004ROUGE},
METEOR~\cite{Denkowski2014Meteor}, and CIDEr~\cite{Vedantam2015CIDEr}. Our
results display a strong correlation between all the metrics---a method that
performs well on one metric tends to perform well on them all. Of particular
interest is CIDEr since it uses Term Frequency Inverse Document Frequency
(TF-IDF) to put more importance on less common words such as entity names. This
makes CIDEr particularly well suited for evaluating news captions where
uncommon words tend to be vitally important, e.g. people's names.

Table \ref{tab:results-names} further reports metrics on the entities. In
particular, we show the precision and recall of all proper nouns and new proper
nouns. We define a proper noun to be new if it has never appeared in any
training caption or article text. This is in contrast to the rare proper noun
metrics reported in the main paper, which are proper nouns that are not present
in any training caption but might have appeared inside a training article
context.

The three rightmost columns of Table \ref{tab:results-names} show the
linguistic quality metrics, including caption length (CL), type-token ratio
(TTR)~\cite{Templin1957CertainLS}, and Flesch readability ease
(FRE)~\cite{Flesch1948,Kincaid1975DerivationON}. The TTR is measured as
\begin{IEEEeqnarray}{lCl}
   \text{TTR} &=& \dfrac{U}{W}
\end{IEEEeqnarray}
where $U$ is the number of unique words and $W$ is the total number of words
in the caption. FRE is measured as
\begin{IEEEeqnarray}{lCl}
   \text{FRE} &=& 206.835-1.015\left({\frac {W}{S}}\right)-84.6\left({\frac {B}{W}}\right)
\end{IEEEeqnarray}
where $W$ is the number of words, $S$ is the number of sentences, and $B$ is
the number of syllables in the caption.

The higher TTR corresponds to a higher vocabulary variation in the text, while
a higher FRE indicates that the text uses simpler words and thus is easier to
read. Overall our models produce captions that are closer in length to the
ground truths than the previous state of the art \textit{Biten}
\cite{Biten2019GoodNews}. Moreover, our captions exhibit a level of language
complexity (as measured by Flesch score) that is closer to the ground truths.
However, there is still a gap in TTR, Flesch, and length, between captions
generated by our model and the human-written ground-truth captions.

\begin{figure*}[p]
   \begin{center}
      \includegraphics[width=\linewidth]{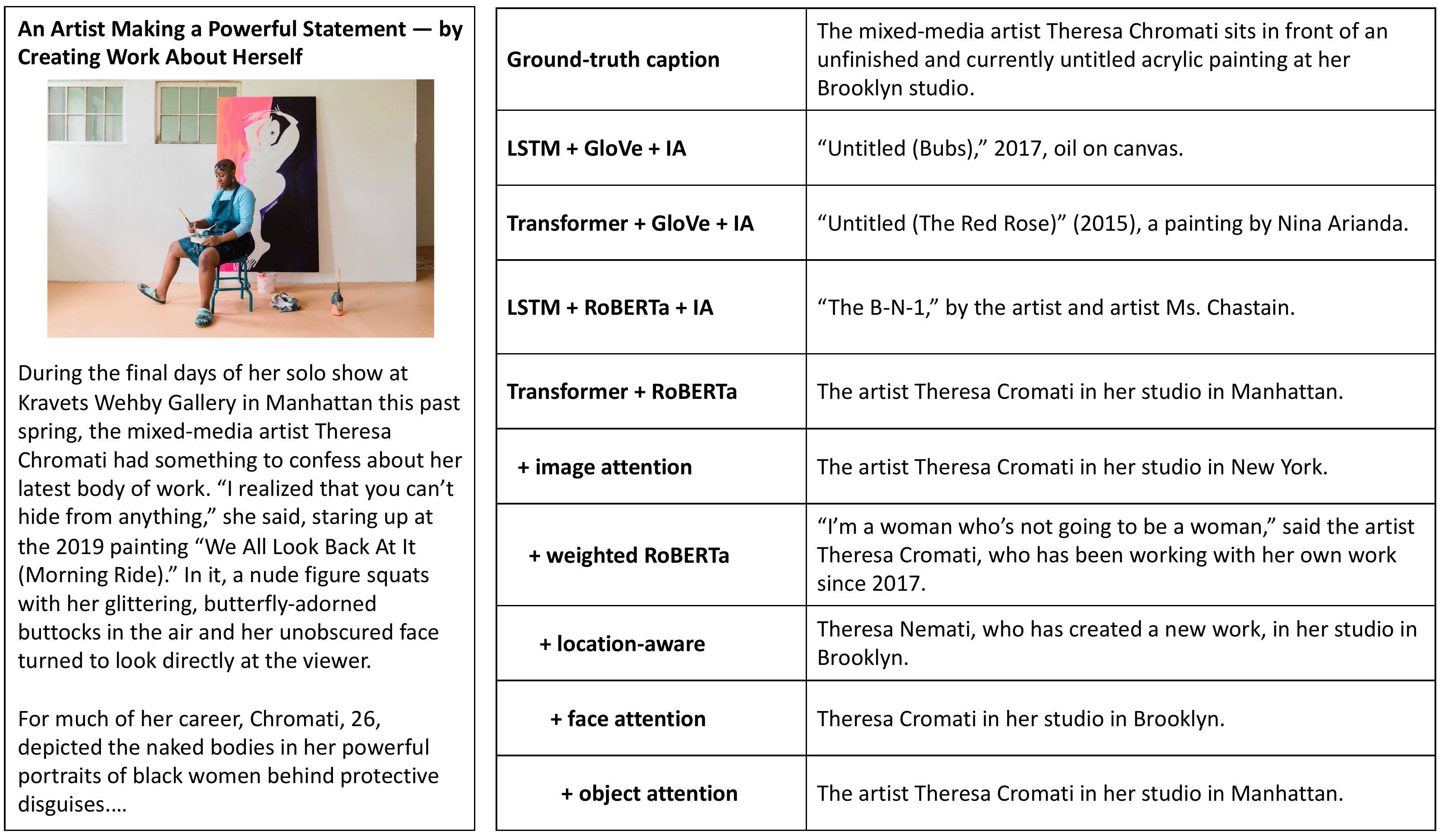}
   \end{center}
   \caption{An example article (left) and the corresponding news captions
      (right) from the NYTimes800k test set. The name ``Chromati" has never
      appeared in the training data, and none of the models can spell the
      artist's name correctly. They all miss the letter ``h'' in her name.
      Captions from models that use an LSTM or GloVe contain made-up names for
      both the painting and the artist. Finally the model that has no access to
      the image, \textit{Transformer + RoBERTa}, still guesses correctly that
      the image is about the artist being in her studio. This shows that
      NYTimes article images can have a predictable theme.}
   \label{fig:chromati}
\end{figure*}

\begin{figure*}[p]
   \begin{center}
      \includegraphics[width=\linewidth]{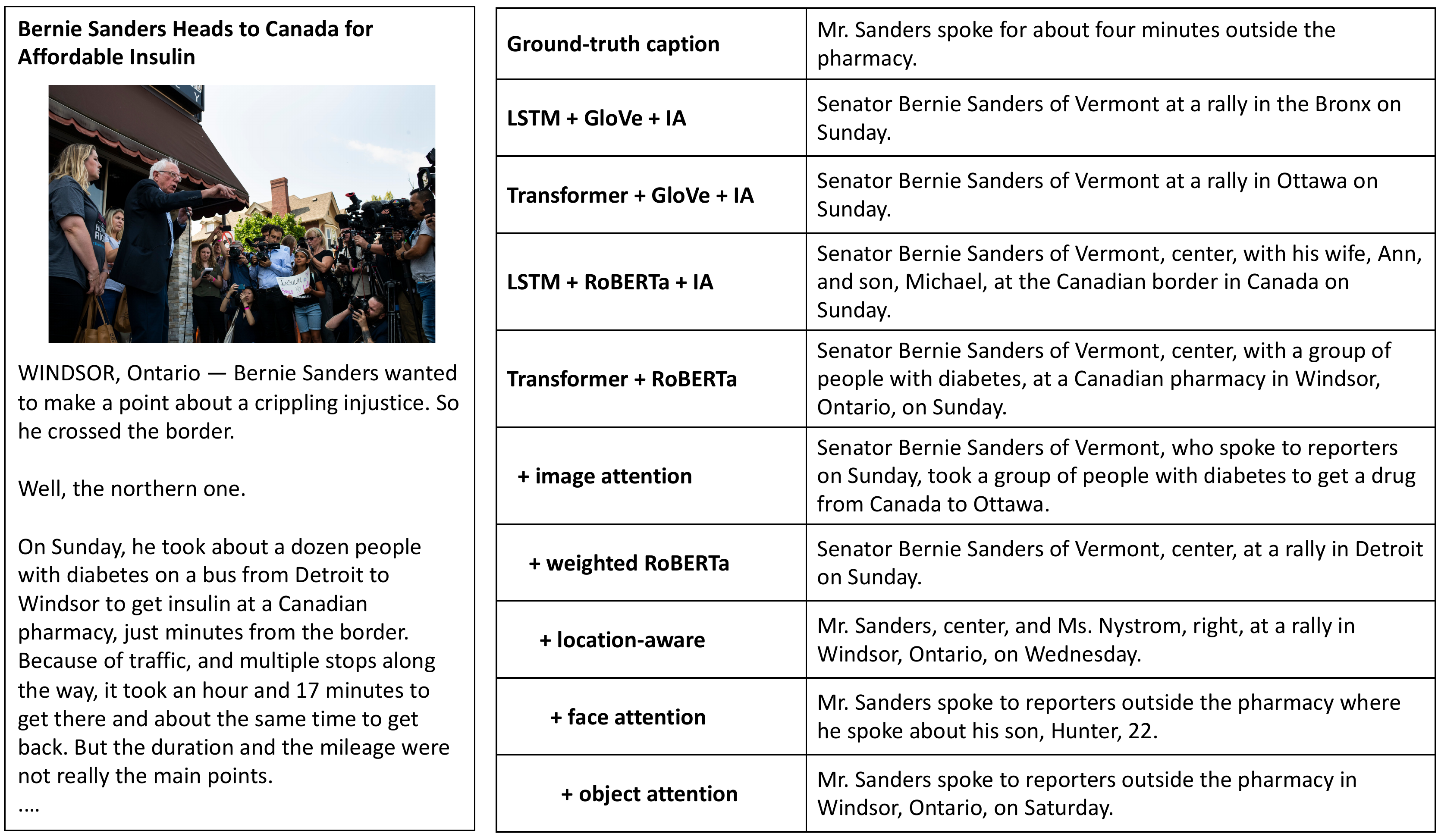}
   \end{center}
   \caption{An example article (left) and the corresponding news captions
      (right) from the NYTimes800k test set. The model that has no access to
      the image, \textit{Transformer + RoBERTa}, is correct in predicting that
      the image is about Bernie Sanders. However it guesses that he is with a
      group of people with diabetes, which is not correct but is sensible given
      the article content. Some of the models manage to override the strong
      prior that he is at a rally (which is what many of Bernie Sanders images
      in the training set are about) and correctly say that he is outside a
      pharmacy. The caption from the model with object attention is the most
      accurate because it generates all three entities correctly: Windsor in
      Ontario, the reporters, and the pharmacy. }
   \label{fig:sanders}
\end{figure*}

Finally Figure~\ref{fig:chromati} and Figure~\ref{fig:sanders} show two
further set of generated captions.

\begin{table*}[p]
   \caption {BLEU, ROUGE, METEOR, and CIDEr metrics on the GoodNews and
      NYTimes800k datasets.}

   \label{tab:results}
   \centering
   \begin{tabularx}{\textwidth}{llXXXXXXX}
      \toprule
       &                                               & BLEU-1 & BLEU-2 & BLEU-3 & BLEU-4 & ROUGE & METEOR & CIDEr \\
      \midrule
      \multirow{10}{*}{\rotatebox[origin=c]{90}{GoodNews}}
       & Biten (Avg + CtxIns)~\cite{Biten2019GoodNews} & 9.04   & 3.66   & 1.71   & 0.89   & 12.2  & 4.37   & 13.1  \\
       & Biten (TBB + AttIns)~\cite{Biten2019GoodNews} & 8.10   & 3.26   & 1.48   & 0.76   & 12.2  & 4.17   & 12.7  \\
       \cmidrule{2-9}

       & LSTM + GloVe + IA              & 14.1   & 6.50   & 3.36   & 1.97   & 13.6  & 5.54   & 13.9  \\
        & Transformer + GloVe + IA         & 18.8   & 9.72   & 5.55   & 3.48   & 17.0  & 7.63   & 25.2  \\
        & LSTM + RoBERTa + IA      & 18.0   & 9.54   & 5.51   & 3.45   & 17.0  & 7.68   & 28.6  \\
      \cmidrule{2-9}

       & Transformer + RoBERTa                        & 19.7   & 11.3   & 6.96   & 4.60   & 18.6  & 8.82   & 40.9  \\
       & \quad + image attention           & 21.6   & 12.7   & 8.09   & 5.45   & 20.7  & 9.74   & 48.5  \\
       & \quad\quad + weighted RoBERTa                 & 22.3   & 13.4   & 8.72   & 6.0    & 21.2  & 10.1   & 53.1  \\
       & \quad\quad\quad + face attention              & \textbf{22.4}   & \textbf{13.5}   & 8.77   & \textbf{6.05}   & \textbf{21.4}  & 10.2   & \textbf{54.3}  \\
       & \quad\quad\quad\quad + object attention       & \textbf{22.4}   & \textbf{13.5}   & \textbf{8.80}   & \textbf{6.05}   & \textbf{21.4}  & \textbf{10.3}   & 53.8  \\
      \midrule
      \midrule \multirow{9}{*}{\rotatebox[origin=c]{90}{NYTimes800k}}

      & LSTM + GloVe + IA              & 13.4   & 6.0    & 3.06   & 1.77   & 13.1  & 5.34   & 12.1  \\
      & Transformer + GloVe + IA         & 16.8   & 8.28   & 4.56   & 2.75   & 15.9  & 6.94   & 20.3  \\
      & LSTM + RoBERTa + IA      & 17.0   & 8.92   & 5.19   & 3.29   & 16.1  & 7.31   & 24.9  \\
      \cmidrule{2-9}

       & Transformer + RoBERTa                        & 18.2   & 10.2   & 6.37   & 4.26   & 17.3  & 8.14   & 33.9  \\
       & \quad + image attention           & 20.0   & 11.6   & 7.38   & 5.01   & 19.4  & 9.05   & 40.3  \\
       & \quad\quad + weighted RoBERTa                 & 20.9   & 12.5   & 8.18   & 5.75   & 19.9  & 9.56   & 45.1  \\
       & \quad\quad\quad + location-aware              & 21.8   & 13.5   & 8.96   & 6.36   & 21.4  & 10.3   & 52.8  \\
       & \quad\quad\quad\quad + face attention         & \textbf{21.6}   & 13.3   & 8.85   & 6.26   & 21.5  & \textbf{10.3}   & 53.9  \\
       & \quad\quad\quad\quad\quad + object attention  & \textbf{21.6}   & \textbf{13.4}   & \textbf{8.90}   & \textbf{6.30}   & \textbf{21.7}  & \textbf{10.3}   & \textbf{54.4}  \\
      \bottomrule
   \end{tabularx}
\end{table*}

\begin{table*}[t]

   \caption {All proper noun and new proper noun precision (P) \& recall (R) on
      the GoodNews and NYTimes800k datasets. Linguistic measures on the
      generated captions: caption length (CL), type-token ratio (TTR), and
      Flesch readability ease (FRE).}

   \label{tab:results-names}
   \centering
   \begin{tabularx}{\textwidth}{llXXXXXX XXX}
      \toprule
       &                                               & \multicolumn{2}{c}{All proper nouns}
       & \multicolumn{2}{c}{New proper nouns}
       & \multirow{2}{*}{CL}                           & \multirow{2}{*}{TTR}                 & \multirow{2}{*}{FRE}                                    \\
       &                                               & P                                    & R                    & P    & R                         \\
      \midrule
      \multirow{8}{*}{\rotatebox[origin=c]{90}{GoodNews}}
       & Ground truths                                 & --                                   & --                   & --   & --   & 18.1 & 94.9 & 65.4 \\
      \cmidrule{2-9}
       & Biten (Avg + CtxIns)~\cite{Biten2019GoodNews} & 16.5                                 & 12.2                 & 2.70 & 12.0 & 9.89 & 92.2 & 78.3 \\
       & Biten (TBB + AttIns)~\cite{Biten2019GoodNews} & 19.2                                 & 11.0                 & 4.21 & 12.3 & 9.14 & 90.7 & 77.6 \\
       \cmidrule{2-9}

       & LSTM + GloVe + IA              & 16.1                                 & 11.3                 & 0    & 0    & 14.0 & 89.5 & 77.2 \\
        & Transformer + GloVe + IA         & 22.7                                 & 18.4                 & 0    & 0    & 16.0 & 88.4 & 73.9 \\
        & LSTM + RoBERTa + IA      & 25.1                                 & 20.8                 & 1.68 & 7.86 & 15.0 & 89.0 & 75.7 \\
      \cmidrule{2-9}

       & Transformer + RoBERTa                        & 30.7                                 & 26.0                 & 7.69 & 16.4 & 15.1 & 90.0 & 73.0 \\
       & \quad + image attention           & 33.4                                 & 28.0                 & 8.53 & 19.3 & 15.2 & 90.0 & 72.5 \\
       & \quad\quad + weighted RoBERTa                 & 33.9                                 & 29.6                 & \textbf{15.2} & \textbf{24.4} & 15.5 & 90.8 & 71.8 \\
       & \quad\quad\quad + face attention              & 34.3                                 & 29.8                 & 13.6 & 22.2 & 15.4 & 90.8 & 71.8 \\
       & \quad\quad\quad\quad + object attention       & \textbf{34.7}                                 & \textbf{29.9}                 & 13.3 & 23.6 & 15.3 & 90.9 & 72.0 \\
      \midrule
      \midrule
      \multirow{7}{*}{\rotatebox[origin=c]{90}{NYTimes800k}}
       & Ground truths                                 & --                                   & --                   & --   & --   & 18.4 & 94.6 & 63.9 \\
      \cmidrule{2-9}

      & LSTM + GloVe + IA              & 15.8                                 & 12.4                 & 0    & 0    & 13.9 & 88.7 & 76.1 \\
      & Transformer + GloVe + IA         & 21.5                                 & 18.2                 & 0    & 0    & 14.8 & 88.8 & 71.9 \\
      & LSTM + RoBERTa + IA      & 24.1                                 & 21.8                 & 3.28 & 7.18 & 14.8 & 89.3 & 73.3 \\
      \cmidrule{2-9}
       & Transformer + RoBERTa                        & 28.0                                 & 26.0                 & 13.4 & 14.5 & 15.2 & 90.4 & 71.4 \\
       & \quad + image attention           & 31.1                                 & 28.7                 & 15.6 & 17.2 & 15.1 & 90.1 & 71.5 \\
       & \quad\quad + weighted RoBERTa                 & 31.8                                 & 30.5                 & 21.7 & 20.2 & 15.5 & 91.6 & 70.1 \\
       & \quad\quad\quad + location-aware              & 36.4                                 & 34.1                 & 26.3 & \textbf{25.3} & 15.1 & 91.7 & 70.8 \\
       & \quad\quad\quad\quad + face attention         & 36.8                                 & 34.2                 & 26.2 & 24.2 & 14.9 & 91.8 & 70.9 \\
       & \quad\quad\quad\quad\quad + object attention  & \textbf{37.2}                                 & \textbf{34.5}                 & \textbf{26.7} & 25.1 & 14.8 & 91.9 & 71.2 \\
      \bottomrule
   \end{tabularx}
\end{table*}

\begin{table*}[t]

   \caption {Geopolitical entity (GPE), organization (ORG), and date (DATE)
      precision (P) \& recall (R) on the GoodNews and NYTimes800k datasets.}

   \label{tab:gpe-org-date}
   \centering
   \begin{tabularx}{\textwidth}{llXXXXXX}
      \toprule
       &                                               & \multicolumn{2}{c}{GPE} & \multicolumn{2}{c}{ORG} & \multicolumn{2}{c}{DATE}                      \\
       &                                               & P                       & R                       & P                        & R    & P    & R    \\
      \midrule
      \multirow{8}{*}{\rotatebox[origin=c]{90}{GoodNews}}
       & Biten (Avg + CtxIns)~\cite{Biten2019GoodNews} & 12.0                    & 11.5                    & 5.67                     & 7.45 & 6.12 & 4.03 \\
       & Biten (TBB + AttIns)~\cite{Biten2019GoodNews} & 12.8                    & 8.41                    & 5.81                     & 7.36 & 5.86 & 4.06 \\
       \cmidrule{2-8}

       & LSTM + GloVe + IA              & 15.6                    & 12.8                    & 14.0                     & 8.58 & 11.0 & 8.20 \\
        & Transformer + GloVe + IA         & 20.8                    & 18.8                    & 16.6                     & 11.8 & 12.0 & 10.1 \\
        & LSTM + RoBERTa + IA      & 20.8                    & 19.2                    & 16.9                     & 12.3 & 13.4 & 10.9 \\
      \cmidrule{2-8}

       & Transformer + RoBERTa                        & 22.6                    & 22.5                    & 20.4                     & 16.3 & 13.8 & 12.6 \\
       & \quad + image attention           & \textbf{25.8}                    & 24.5                    & 21.0                     & 17.3 & 14.4 & 13.0 \\
       & \quad\quad + weighted RoBERTa                 & 25.0                    & 24.2                    & 22.0                     & \textbf{18.7} & 14.3 & 13.1 \\
       & \quad\quad\quad + face attention              & 24.9                    & 24.4                    & 21.6                     & 18.5 & 14.7 & \textbf{13.3} \\
       & \quad\quad\quad\quad + object attention       & 25.6                    & \textbf{24.7}                    & \textbf{22.4}                     & \textbf{18.7} & \textbf{15.1} & \textbf{13.3} \\
      \midrule
      \midrule
      \multirow{7}{*}{\rotatebox[origin=c]{90}{NYTimes800k}}
      & LSTM + GloVe + IA              & 16.0                    & 14.7                    & 8.60                     & 4.89 & 11.3 & 8.31 \\
      & Transformer + GloVe + IA         & 19.1                    & 21.8                    & 12.1                     & 7.95 & 11.3 & 10.1 \\
      & LSTM + RoBERTa + IA      & 20.2                    & 22.2                    & 13.1                     & 8.95 & 11.8 & 11.1 \\
      \cmidrule{2-8}
       & Transformer + RoBERTa                        & 21.4                    & 25.4                    & 15.8                     & 12.2 & 12.0 & 12.5 \\
       & \quad + image attention           & 23.9                    & 27.3                    & 17.6                     & 13.6 & 12.8 & 13.2 \\
       & \quad\quad + weighted RoBERTa                 & 24.2                    & 28.2                    & 19.2                     & 15.6 & 13.9 & 14.3 \\
       & \quad\quad\quad + location-aware              & 26.8                    & 30.1                    & 20.9                     & \textbf{17.3} & \textbf{14.1} & \textbf{14.1} \\
       & \quad\quad\quad\quad + face attention         & \textbf{26.9}                    & \textbf{30.6}                    & 20.7                     & 16.5 & 13.9 & \textbf{14.1} \\
       & \quad\quad\quad\quad\quad + object attention  & 26.8                    & \textbf{30.6}                    & \textbf{21.9}                     & 17.2 & 13.7 & 13.8 \\

      \bottomrule
   \end{tabularx}
\end{table*}

\clearpage

\end{document}